\documentclass[11pt,a4paper]{article}
\usepackage[hyperref]{naaclhlt2019}
\usepackage{latexsym}
\usepackage{times,latexsym}
\usepackage{url}
\usepackage[T1]{fontenc}
\usepackage{times}
\usepackage{helvet}
\usepackage{courier}
\usepackage{times}
\usepackage{latexsym}
\usepackage{graphicx}
\usepackage{amsmath}
\usepackage{graphicx}
\usepackage{array}
\usepackage{makecell}
\usepackage{fontawesome}

\usepackage{url}

\aclfinalcopy 

\title{Multi-Label Transfer Learning for Multi-Relational Semantic Similarity}

\author{Li Zhang \and Steven R. Wilson \and Rada Mihalcea \\
Computer Science and Engineering \\
  University of Michigan \\
  {\tt \{zharry,steverw,mihalcea\}@umich.edu} }
\date{}

\begin{document}
\maketitle
\begin{abstract}
    Multi-relational semantic similarity datasets define the semantic relations between two short texts in multiple ways, e.g., similarity, relatedness, and so on. Yet, all the systems to date designed to capture such relations target one relation at a time. We propose a multi-label transfer learning approach based on LSTM to make predictions for several relations simultaneously and aggregate the losses to update the parameters. This multi-label regression approach jointly learns the information provided by the multiple relations, rather than treating them as separate tasks. Not only does this approach outperform the single-task approach and the traditional multi-task learning approach, but it also achieves state-of-the-art performance on all but one relation of the Human Activity Phrase dataset.
\end{abstract}

\section{Introduction}

Semantic similarity, or relating short texts or sentences\footnote{In this work, we do not consider word level similarity.} in a semantic space -- be those phrases, sentences or short paragraphs -- is a task that requires systems to determine the degree of equivalence between the underlying semantics of the two sentences.
Although relatively easy for humans, this task remains one of the most difficult natural language understanding problems. 
The task has been receiving significant interest from the research community. For instance, from 2012 to 2017, the International Workshop on Semantic Evaluation (SemEval) has been holding the Semantic Textual Similarity (STS) shared tasks  \cite{Agirre12,Agirre13,agirre2015semeval,agirre2016semeval,cer2017semeval}, dedicated to tackling this problem, with close to 100 team submissions each year.

In some semantic similarity datasets, an example consists of a sentence pair and a single annotated similarity score, while in others, each pair comes with multiple annotations. We refer to the latter as \textit{multi-relational semantic similarity} tasks. The inclusion of multiple annotations per example is motivated by the fact that there can be different relations, namely different types of similarity between two sentences. So far, these relations have been treated as separate tasks, where a model trains and tests on one relation at a time while ignoring the rest. However, we hypothesize that each relation may contain useful information about the others, and training on only one relation inevitably neglects some relevant information. Thus, training jointly on multiple relations may improve performance on one or more relations. 

We propose a joint {\bf multi-label} transfer learning setting based on LSTM, and show that it can be an effective solution for the multi-relational semantic similarity tasks.  Due to the small size of multi-relational semantic similarity datasets and the recent success of LSTM-based sentence representations \cite{Wieting17pushing,Conneau17}, the model is pre-trained on a large corpus and transfer learning is applied using fine-tuning. In our setting, the network is jointly trained on multiple relations by outputting multiple predictions (one for each relation) and aggregating the losses during back-propagation. This is different from the traditional multi-task learning setting where the model makes one prediction at a time, switching between the tasks. We treat the multi-task setting and the single-task setting (i.e., where a separate model is learned for each relation) as baselines, and show that the multi-label setting outperforms them in many cases, achieving state-of-the-art performance on all but one relation of the Human Activity Phrase dataset \cite{Wilson17}. 

In addition to success on multi-relational semantic similarity tasks, the multi-label transfer learning setting that we propose can easily be paired with other neural network architectures and applied to any dataset with multiple annotations available for each training instance.

\begin{figure}[t!]
	\includegraphics[width=0.47\textwidth]{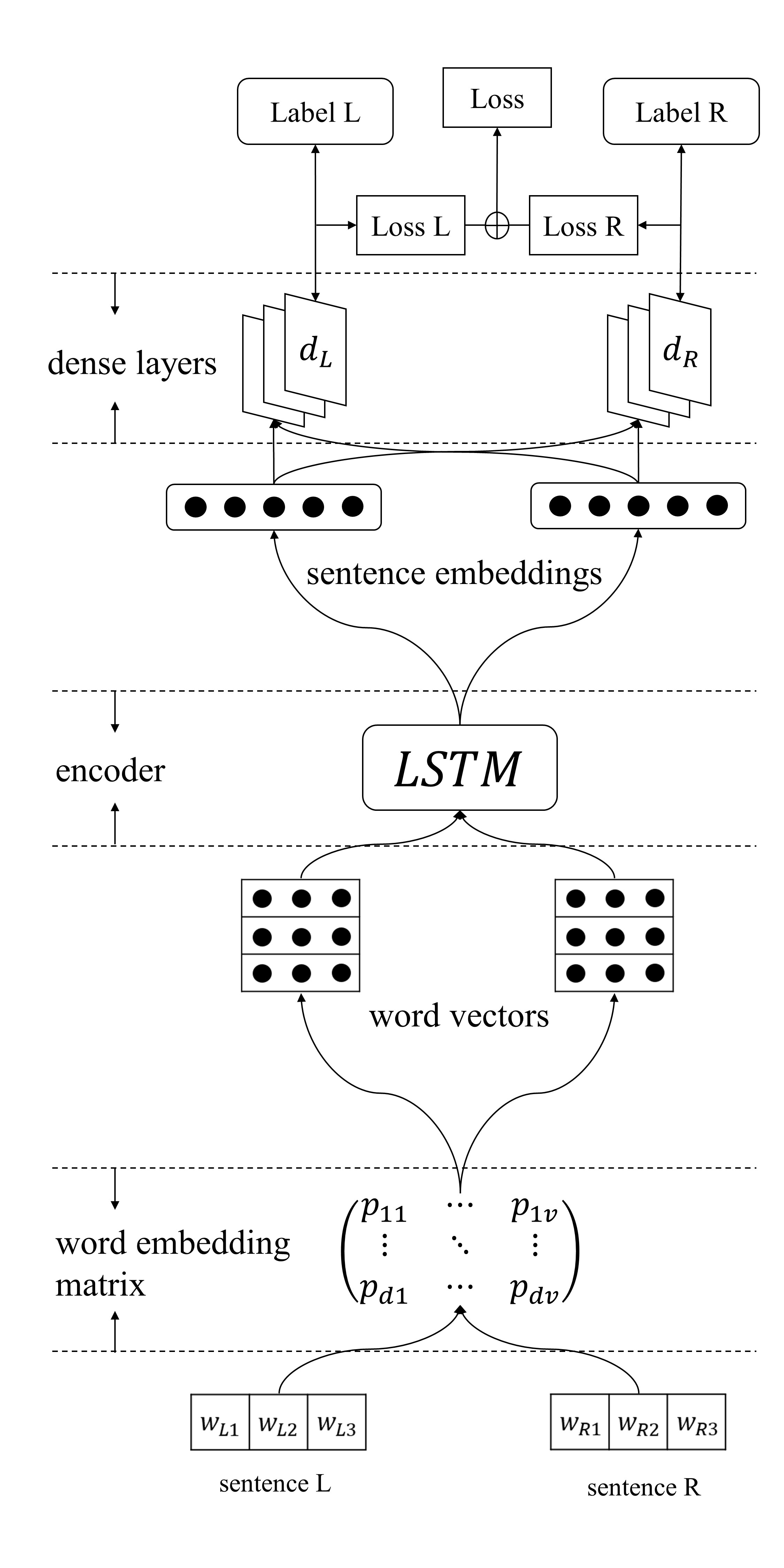}
	\caption{\label{architecture} Overview of the multi-label architecture.}
\end{figure}

\section{Multi-Label Transfer Learning}

We introduce a multi-label transfer learning setting by modifying the architecture of the LSTM-based sentence encoder, specifically designed for multi-relational semantic similarity tasks. 

\subsection{Architecture}
We employ the ``hard-parameter sharing'' setting \cite{caruana1998multitask}, where some hidden layers are shared across multiple tasks while each task has its own specific output layer. As shown in Figure~\ref{architecture}, using an example of a semantic similarity dataset with two relations, sentence L and sentence R in a pair are first mapped to word vector sequences and then encoded as sentence embeddings. Up to this step, the choice of the word embedding matrix and sentence encoder is flexible, and we outline our choice in the sections to follow. For each relation that has been annotated with a ground-truth label, a dedicated output dense layer takes the two sentence embeddings as input and outputs a probability distribution across the range of possible scores. The output dense layers follow the methods of Tai et al. \shortcite{Tai15}.

With two such dense output layers, two losses are calculated, one for each relation. The total loss is calculated as the sum of the two losses for back-propagation which updates all parameters in the end-to-end network. 

\subsection{Model}

We use InferSent \cite{Conneau17} as the sentence encoder due to its outstanding performances reported on various semantic similarity tasks. 

Due to the small sizes of the evaluation datasets, we use the sentence encoder pre-trained on the Stanford Natural Language Inference corpus \cite{Bowman15} and Multi-Genre Natural Language Inference corpus \cite{williams2017broad}, and transfer to the semantic similarity tasks using fine-tuning. In this process, the output layers for multi-label learning discussed above are stacked on top of the InferSent network, forming an end-to-end model for training and testing on semantic similarity tasks.

\subsection{Comparison with Multi-Task Learning}

Neither multi-task nor multi-label learning have been used for multi-relational semantic similarity datasets. For these datasets, either multi-task or multi-label learning can be achieved by treating each relation as a ``task.'' The key differences between the two are the relations involved in each forward-backward pass and the timing of the parameter updates. 

Consider a training step in the two-relation example in Figure~\ref{architecture}: 

A \textbf{multi-task learning} model would pick a batch of sentences pairs, only consider \textit{Label L}, only calculate \textit{Loss L}, and all parameters except those of dense layer $d_R$ are updated. Then, within the same batch,\footnote{In general multi-task learning, a new batch is picked after switching tasks. In multi-relational semantic similarity datasets, each task is a relational label, which shares the same input.} the model would only consider \textit{Label R}, only calculate \textit{Loss R}, and all parameters except those of dense layer $d_L$ are updated.

A \textbf{multi-label learning} model (our model) would pick a batch of sentences pairs, consider both \textit{Label L} and \textit{Label R}, calculate \textit{Loss L} and \textit{Loss R}, aggregate them as the total loss, and update all parameters. 

\section{Experiments}

To show the effectiveness of the multi-label transfer learning setting, we experiment on three semantic similarity datasets with multiple relations annotated, and use one LSTM-based sentence encoder that has been very successful in many downstream tasks. 

\subsection{Datasets} \label{datasets}

We study three semantic similarity datasets with multiple relations with texts of different lengths, spanning phrases, sentences, and short paragraphs.\\

\noindent \textbf{Human Activity Phrase} \cite{Wilson17}: a collection of pairs of phrases regarding human activities, annotated with the following four different relations.
\begin{itemize}
    \item Similarity (SIM): The degree to which the two activity phrases describe the same thing, semantic similarity in a strict sense. Example of high similarity phrases: \textit{to watch a film} and \textit{to see a movie}.
    \item Relatedness (REL): The degree to which the activities are related to one another, a general semantic association between two phrases. Example of strongly related phrases: \textit{to give a gift} and \textit{to receive a present}.
    \item Motivational alignment (MA): The degree to which the activities are (typically) done with similar motivations. Example of phrases with potentially similar motivations: \textit{to eat dinner with family members} and \textit{to visit relatives}.
    \item Perceived actor congruence (PAC): The degree to which the activities are expected to be done by the same type of person. An example of a pair with a high PAC score: \textit{to pack a suitcase} and \textit{to travel to another state}.
\end{itemize}
The phrases are generated, paired and scored on Amazon Mechanical Turk.\footnote{https://www.mturk.com/} The annotated scores range from $0$ to $4$ for SIM, REL and MA, and $-2$ to $2$ for PAC. The evaluation is based on the Spearman's $\rho$ correlation coefficient between the systems' predicted scores and the human annotations. There are 1,000 pairs in the dataset. We also use the supplemental 1,373 pairs from Zhang et al. \shortcite{zhang2018sequential} in which 1,000 pairs are randomly selected for training and the rest are used for development. We then treat the original 1,000 pairs as a held-out test set so that our results are directly comparable with those previously reported.\\

\noindent \textbf{SICK} \cite{Marelli14a,Marelli14b}: the Sentences Involving Compositional Knowledge benchmark, which includes a large number of sentence pairs that are rich in the lexical, syntactic and semantic phenomena. Each pair of sentences is annotated in two dimensions: relatedness and entailment. The relatedness score ranges from $1$ to $5$, and Pearson's $r$ is used for evaluation; the entailment relation is categorical, consisting of entailment, contradiction, and neutral. There are 4439 pairs in the train split, 495 in the trial split used for development and 4906 in the test split. The sentence pairs are generated from image and video caption datasets before being paired up using some algorithm. Due to the lack of human supervision in the process, some sentence pairs display minimal difference in semantic components, making the SICK tasks simpler than the others we study. \\

\noindent \textbf{Typed-Similarity} \cite{Agirre13}: a collection of meta-data describing books, paintings, films, museum objects and archival records taken from Europeana,\footnote{http://www.europeana.eu/} presented as the pilot track in the SemEval 2013 STS shared task. Typically, the items consist of title, subject, description, and so on, describing a cultural heritage item and, sometimes, a thumbnail of the item itself. For the purpose of measuring semantic similarity, we concatenate all the textual entries such as title, creator, subject and description into a short paragraph that is used as input, although the annotations might be informed of the image aspects of the meta-data. Each pair of items is annotated on eight dimensions of similarity: general similarity, author, people involved, time, location, event or action involved, subject and description. There are 750 pairs in the train split, of which we randomly sample 500 for training and 250 for development, and 721 in the test split. Pearson's $r$ is used for evaluation.

\subsection{Baselines}

We compare the multi-label setting with two baselines:
\begin{itemize}
    \item \textbf{Single-task}, where each relation is treated as an individual task. For each relation, a model with only one output dense layer is trained and tested, ignoring the annotations of all other relations. 
    \item \textbf{Multi-task}, where only one relation is involved during each round of feed-forward and back-propagation. 
\end{itemize}

\subsection{Experimental Details}

In each experiment, we use stochastic gradient descent and a batch size of 16. We tune the learning rate over $\{0.1, 0.5, 1, 5\}$ and number of epochs over $\{10, 20\}$. For each dataset discussed above, we tune these hyperparameters on the development set. All other hyperparameters maintain their values from the original code.\footnote{https://github.com/facebookresearch/InferSent} In the single-task setting, the model is trained and tested on each relation, ignoring the annotations of other relations. In the multi-task settings, the model is trained and tested on all the relations in a dataset. In the multi-task setting, relations are presented to the model in the order they are listed in the result tables within each batch.

\section{Evaluation}

The results are shown in Tables~\ref{typed-results}, \ref{activity-results} and \ref{sick-results}. For every experiment (represented by a cell in the tables), 30 runs with different random seeds are recorded and the average is reported. For each relation (each column in the tables), let the true mean performance of multi-label learning, single-task baseline and multi-task baseline be $\mu_\text{MLL}$, $\mu_\text{single}$, $\mu_\text{MTL}$, respectively. Two one-sided Student's t-tests are conducted to test if multi-label learning outperforms the baselines for that relation. The significance level is chosen to be $0.05$. A down-arrow $\downarrow$ indicates that our proposed multi-label learning underperforms a baseline, while an up-arrow $\uparrow$ indicates that our proposed multi-label learning outperforms a baseline.

\section{Discussion}

\subsection{Results}

\begin{table*}[t!]
	\begin{center}	
		\begin{tabular}{c|c|c|c|c|c|c|c|c}
			\hline & \bf general  & \bf author  & \bf people  & \bf time & \bf location  & \bf event  & \bf subject  & \bf description  \\ \hline
			 MLL & .744 & .721 & .640 & .713 & .751 & .611 & .697 & .737 \\
			 Single & .750$\downarrow$ & .690$\uparrow$ & .619$\uparrow$ & .712 & .744$\uparrow$ & .606$\uparrow$ & .694$\uparrow$ & .718$\uparrow$ \\
			 MTL & .718$\uparrow$ & .689$\uparrow$ & .611$\uparrow$ & .697$\uparrow$ & .723$\uparrow$ & .579$\uparrow$ & .669$\uparrow$ & .714$\uparrow$ \\
			 \hline
		\end{tabular}
	\end{center}
	\caption{\label{typed-results} The performance in Pearson's $r$ on the Typed-Similarity dataset, in accordance with the specification of the dataset to allow for direct comparison with previous results. The results of single task and multi-task learning (MTL) are followed by $\uparrow$ if it is statically significantly lower than those of multi-label learning (MLL), and they are followed by $\downarrow$ otherwise.}
\end{table*}

\begin{table}[t!]
	\begin{center}	
		\begin{tabular}{c|c|c|c|c}
			\hline \bf   & \bf SIM  & \bf REL  & \bf MA & \bf PAC  \\ \hline
			MLL & .720 & .721 & .682 & .557 \\
			Single & .719 & .717$\uparrow$ & .682 & .555 \\
			MTL & .683$\uparrow$ & .686$\uparrow$ & .651$\uparrow$ & .515$\uparrow$ \\
			\hline
		\end{tabular}
	\end{center}
	\caption{\label{activity-results} The performance in Spearman's $\rho$ on the Human Activity Phrase dataset. }
    \end{table}

\begin{table}[t!]
	\begin{center}	
		\begin{tabular}{c|c|c}
			\hline \bf   & \bf Relatedness  & \bf Entailment  \\ \hline
			MLL & .882 & 86.7 \\
			Single & .874$\uparrow$ & 86.4$\uparrow$ \\
			MTL & .871$\uparrow$ & 86.2$\uparrow$ \\
			\hline
		\end{tabular}
	\end{center}
	\caption{\label{sick-results} The performance in Pearson's $r$ on the SICK dataset, in accordance with the specification of the dataset to allow for direct comparison with previous results.}
\end{table}

For the Human Activity Phrase dataset, the single-task setting already achieves state-of-the-art performances on SIM, REL and PAC relations, surpassing the previous best results reported by Zhang et al. \shortcite{zhang2018sequential}, which achieved Spearman's correlation coefficient of .710 in SIM, .715 in REL, .690 in MA and .549 in PAC. This approach is based on fine-tuning a bi-directional LSTM with average-pooling pre-trained on translated texts \cite{Wieting17pushing}.  Using multi-label learning, our model is able to gain a statistically significant improvement in the performance of REL compared to the single-task setting, while maintaining performance for the other relations. The traditional multi-task setting, however, performs significantly worse than the other settings. 

For the entailment task on the SICK dataset, our multi-label setting outperforms the single-task baseline and the previous best results of InferSent. These best results consisted of  an accuracy of 86.3\% achieved  using a logistic regression classifier and sentence embeddings generated by pre-trained InferSent as features \cite{Conneau17}. In the relatedness task, this setting achieved a Pearson's correlation coefficient of .885, which even our our multi-label setting is unable to beat. However, the multi-label setting does have a statistically significant performance gain compared to the single-task setting in the relatedness task, while the traditional multi-task setting underperforms the other settings. 

For the Typed-Similarity dataset, the previous best results are achieved using rich feature engineering without the use of sentence embeddings, with a different scoring scheme for each relation \cite{agirre2013ubc_uos}. While this method yielded better results than all of the transfer learning approaches we compare, it should be noted that this approach is specific to tackling this dataset, unlike the transfer learning settings that are generalizable to other scenarios. One potential reason for the discrepancy in performance is that some relations such as time, people involved, or events may be easily or sometimes trivially captured by information retrieval techniques such as named entity recognition. Using sentence embeddings and transfer learning for all the relations, though simpler, may face greater challenge in the relations mentioned above. Among the three transfer learning approaches, our multi-label setting is still superior, outperforming the single-task setting in over half of the relations, and outperforming the multi-task setting in all relations. 

\subsection{Empirical Recommendation}

While our results above show that multi-label learning is almost always the most effective way to transfer sentence embeddings in multi-relational semantic similarity tasks, in some situations simply training with one relation might yield better performance (such as the general similarity relation in the Typed-Similarity dataset). This suggests that the choice of multi-label learning or single-task learning can be tuned as a hyperparameter empirically for the optimal performance on a task. 

\subsection{Other Considerations and Discussions}

In the multi-label setting, we calculate the total loss by summing the loss from each dimension. We also explore weighting the loss from each dimension by factors of 2, 5 and 10, but doing so hurts the performance for all dimensions. 

In the multi-task setting, we attempt different ordering of the dimensions when presenting them to the model within a batch of examples, but the difference in performance is not statistically significant. Furthermore, the multi-task setting takes about $n$ times longer to train than the multi-label setting, where $n$ is number of dimensions of annotations.

\section{Conclusions}

We introduced a multi-label transfer learning setting designed specifically for semantic similarity tasks with multiple relations annotations. By experimenting with a variety of relations in three datasets, we showed that the multi-label setting can outperform single-task and traditional multi-task settings in many cases. 

Future work includes exploring the performance of this setting with other sentence encoders, as well as multi-label datasets outside of the domain of semantic similarity. This may include NLP datasets annotated with author information for multiple dimensions, or computer vision datasets with multiple annotations for scenes. 

\section*{Acknowledgments}
This material is based in part upon work supported by the Michigan Institute for Data Science, by the John Templeton Foundation (grant \#61156), by the National Science Foundation (grant \#1815291), and by DARPA (grant \#HR001117S0026-AIDA-FP-045). Any opinions, findings, and conclusions or recommendations expressed in this material are those of the author and do not necessarily reflect the views of the Michigan Institute for Data Science, the John Templeton Foundation, the National Science Foundation,  or DARPA. 



\bibliography{naaclhlt2019}
\bibliographystyle{acl_natbib}

\end{document}